\documentclass[sigconf]{acmart}

\usepackage{booktabs} 
\usepackage{subcaption,graphicx}
\usepackage{appendix}

\setcopyright{acmcopyright}
\usepackage{todonotes,url}

\acmDOI{	arXiv:1705.09451}

\acmConference[KDD Workshop on ML meets Fashion]{KDD}{August 2017}{Canada} 
\acmYear{2017}
\copyrightyear{2017}

\usepackage{navigator}

\begin{document}

\title{Algorithmic clothing: hybrid recommendation, from street-style-to-shop}


\author{Y Qian}
\author{P Giaccone}
\author{M Sasdelli}
\author{E Vazquez}
\affiliation{%
  \institution{Cortexica Vision Systems Limited}
  \streetaddress{Capital Tower -- 91 Waterloo Road}
  \city{London} 
  \state{United Kingdom} 
  \postcode{SE1 8RT}
}

\author{B Sengupta}
\authornote{corresponding author}
\affiliation{%
\institution{Dept. of Engineering, University of Cambridge}
\institution{Dept. of Bio-engineering, Imperial College London}
\institution{Cortexica Vision Systems Limited}
\state{United Kingdom}
}
\email{bs573@cam.ac.uk}

\renewcommand{\shortauthors}{Qian and Sengupta}

\begin{abstract} 
In this paper we detail Cortexica's (\url{https://www.cortexica.com/}) recommendation framework -- particularly, we describe how a hybrid visual recommender system can be created by combining conditional random fields for segmentation and deep neural networks for object localisation and feature representation. The recommendation system that is built after localisation, segmentation and classification has two properties -- first, it is knowledge based in the sense that it learns pairwise preference/occurrence matrix by utilising knowledge from experts (images from fashion blogs) and second, it is  content-based as it utilises a deep learning based framework for learning feature representation. Such a construct is especially useful when there is a scarcity of user preference data, that forms the foundation of many collaborative recommendation algorithms.
\end{abstract} 

%
%




\maketitle

\section{Introduction}
\label{sec:intro}

Algorithmic clothing frameworks have been routinely operationalized using recommender systems that utilise deep convolutional networks for feature representation and prediction \cite{Sengupta2017}. Such a content-based image retrieval amounts to recommending items to users who might have similar styles (contemporary vs. retro, etc.), like similar patterns (stripes vs. polka, etc.) or colours. Often time, frameworks rely on recommending not a single item but a pair (or triplets, etc.) of products. For the fashion vertical, this would mean recommending what trousers to wear with a given shirt, for example. Yet, service providers often do not have access to a consumer's behaviour -- be it click-through rate, prior purchases, etc. This means the recommender system generally has a cold-start problem, resulting in incorrect recommendations. This is a problem intrinsic to many collaborative recommendation algorithms which are built with the \textit{a priori} assumption of the availability of a user preference database. 

In this paper, we describe a scalable method to recommend fashion inventory (tops, trousers, etc.) wherein user preferences are learnt from experts (oracles), that we accumulate using images from fashion blogs. We call them `street-style images'. Utilising deep neural networks enable us to parse such images into a high dimensional feature representation that allow us to recommend a pair from a high-dimensional user preference tensor that can be constrained to exist in a retailer's database. Whilst the database presented in this paper has a 2-D form, utilising a deep neural network enables us to learn an n-D tensor. This helps in recommending multiple items that have a structure -- for example, to predict not only whether a trouser fits well with a shirt but also what allied accessories like belts, cuffs, etc. can be worn.

In section \ref{sec:methods}, we detail the technical infrastructure that allows us to produce recommendations starting from a single image. We start by (a) segmenting and localising the object of interest and (b) designing a knowledge base that constrains the recommendation from the knowledge derived from street-style images to only those commodities that are available from a retailer's database.

\section{Methods} 
\label{sec:methods}

Recommending an item that is suitable to wear with yet another item involves the following steps:

\begin{itemize}
\item Localisation/segmentation of garments from street-style images (oracle)
\item Generation of association between a pair of garment, i.e., determine which items in each image are being worn by the same person
\item Construction of a joint distribution (co-occurrence matrix) based on either visual features from street-style images or items from a vendor's inventory
\item Produce recommendation by using: (a) colour, (b) pattern or (c) a street-style oracle under a content-based retrieval framework
\end{itemize}

\begin{figure}
 \centering \includegraphics[width=0.5\textwidth, height=2in]
 {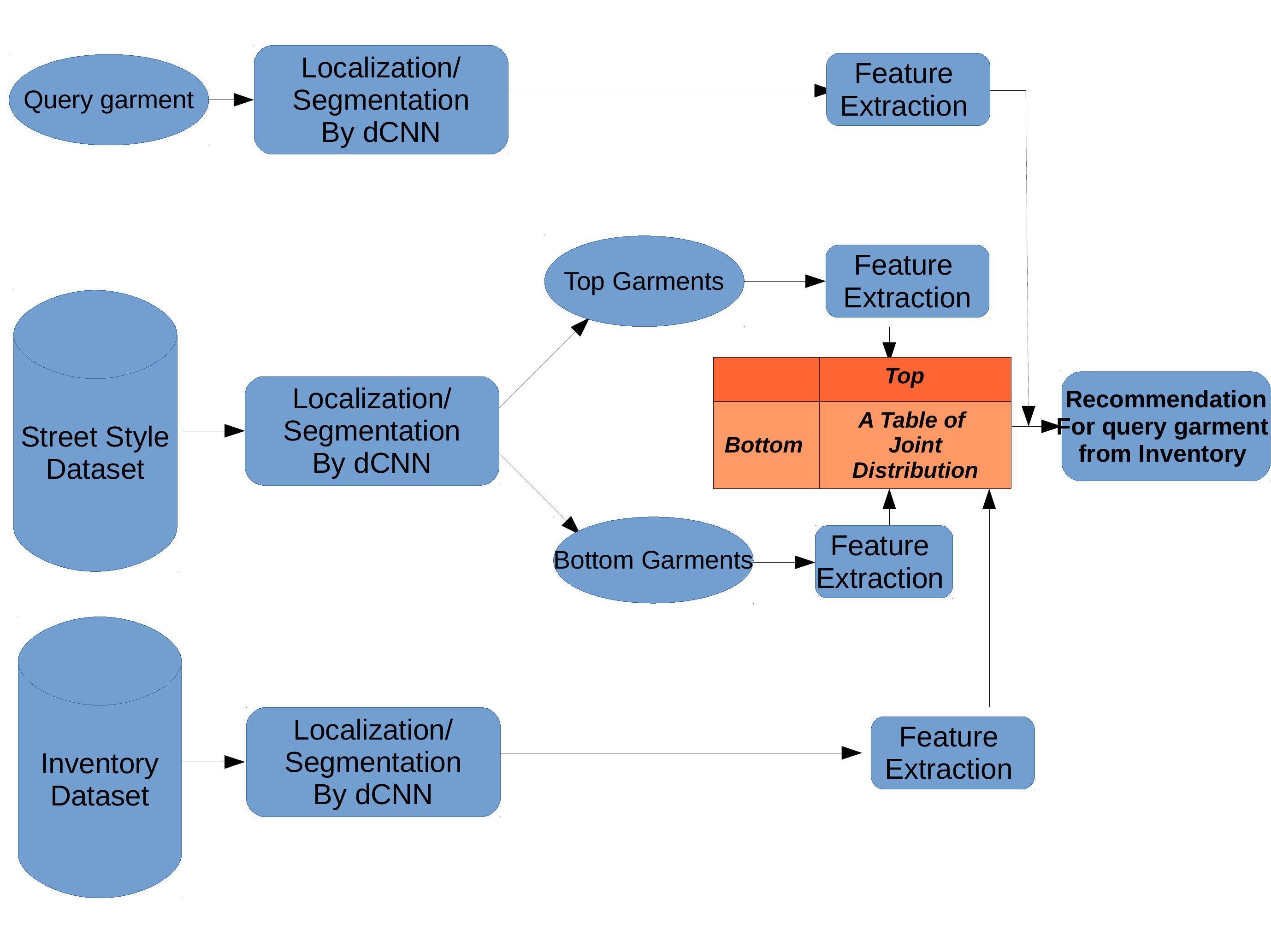}
 \caption{Framework for Fashion Recommendation System (FRS). Garments in street-style images are first cropped/segmented by using a dCNN (deep Convolutional Neural Network). An association of a pair of garment (top with bottom) is then generated, followed by visual feature extraction; eventually, a joint feature distribution is constructed from street-style images. Joint distribution can also be generated from inventory dataset by utilizing content based retrieval framework and finessing such knowledge by the street-style knowledgebase. }
 \label{fig_FRS}
\end{figure}

In the next sections, we will describe the steps comprising a recommendation engine (Figure \ref{fig_FRS}) starting with a description of deep convolutional neural networks (dCNN). A pre-trained VGG-16 model \citep{simonyanZ14a} on ImageNet dataset is used as a base model to fine tune our fashion segmentation, localisation and pattern classification models.

\subsection*{Segmentation using a dCNN} 
\label{sec:segmentation}

\begin{figure}
 \centering \includegraphics[width=0.5\textwidth, height=2.5in]
 {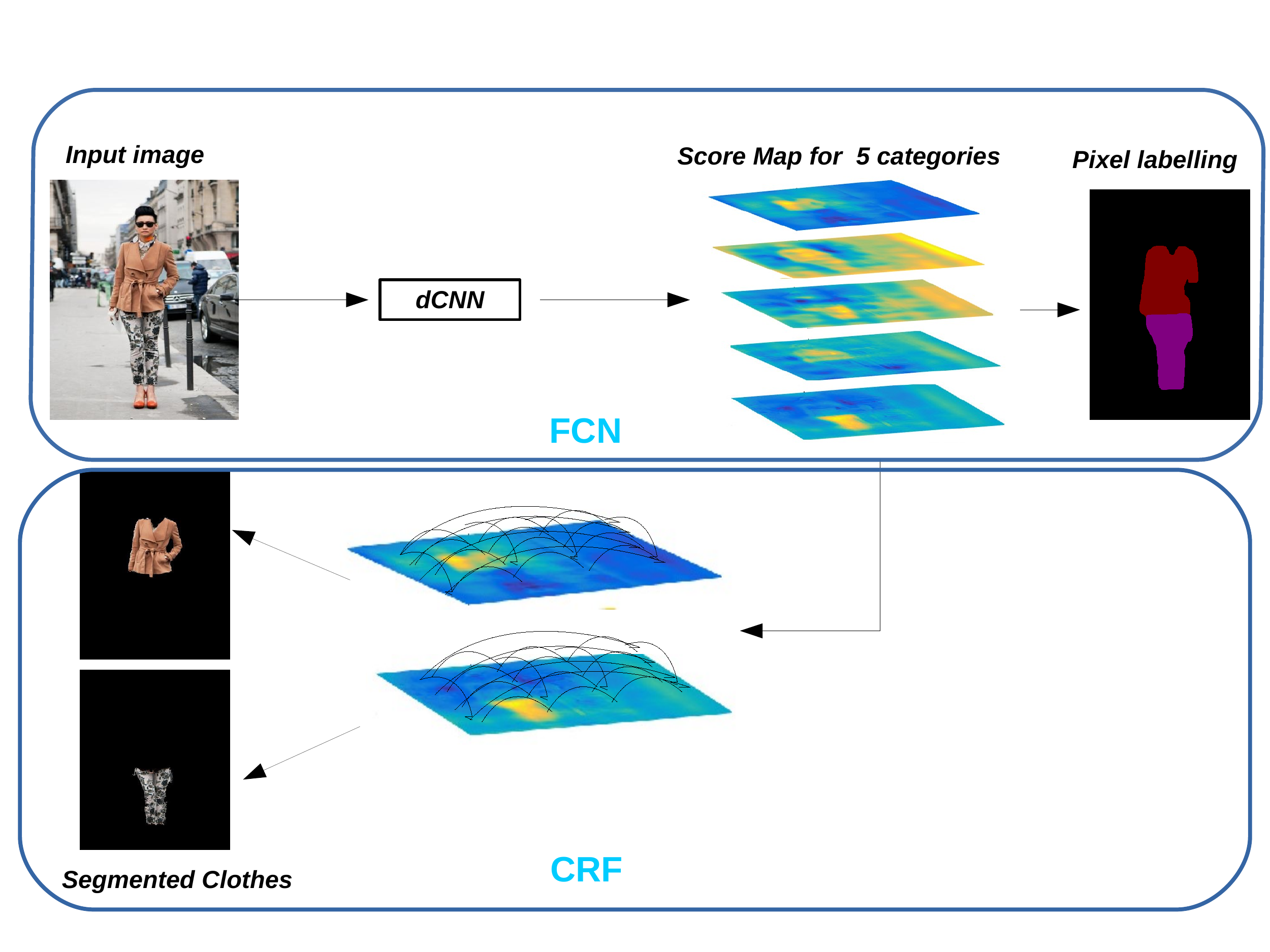}
 \caption{Framework for deep segmentation. Top: pixelwise prediction by using a FCN; the output are the score maps for different categories that can be used to label pixels by finding a category with the maximum value. Bottom: a fully connected CRF is applied to further refine segmented results by combining score maps (pixel label assignment probability) with fully-connected graph on all pairs of original image pixels.}
 \label{fig_DeepSegmentation}
\end{figure}

 Images are segmented using a dCNN framework so that the images could be partitioned into groups of unique objects. Specifically, in order to remove background effect for the dominant colour generation, each clothing item in street-style images is segmented by using an FCN (Fully Convolutional Network) followed by a CRF (Conditional Random Fields), as shown in Figure \ref{fig_DeepSegmentation}. 

Two methods -- FCN \citep{Long15semantic} and DeepLab \citep{chen15semantic} are evaluated on 10K street-style images by annotating individual items with a mask (in Section \ref{sec:DatasetSeg}). Both methods first convert the pre-trained dCNN classifier by replacing the last fully connected layers by fully convolutional layers; this produces coarse output maps. For pixel-wise prediction, upsampling and concatenating the scores from intermediate feature maps are applied to connect these coarse outputs back to the pixels. DeepLab speeds up segmentation of dCNN feature maps by using the `atrous' (with holes) algorithm \cite{Mallat2008}. Instead of deconvolutional layers, the atrous algorithm is applied in the layer that follows the last two max-pooling layers. This is done by introducing zeros to convolutional filters to increase their length. This can control the Field-of-View (FOV)  of the  models by  adjusting the  input stride, without increasing the number of parameters or the amount of computation. Additionally, atrous spatial pyramid pooling (ASPP) is employed in DeepLab to encode objects as well as image context at multiple scales. After coupling with dCNN based pixel-wise prediction (blob-like coarse segmentation), a fully-connected pairwise Conditional Random Field (CRF) \cite{Krahenbuhl11CRF} is applied to model the fine edge details. This is done by coupling neighbouring nodes to assign the same label to spatially proximal pixels.  

\subsection*{Localisation using a dCNN}
\label{sec:localization}

Due to complicated backgrounds in street-style images, object detection is applied to localize the clothing items from the cluttered background. The detected garments are used as a query to find similar items from same class inventory images;  for details on the deep learning architecture  refer to \cite{Sengupta2017}, for feature encoding refer to \cite{Sengupta2017a} and for similarity measurements please see \cite{Qian2017}. Additionally, detected items are classified according to different texture patterns. With the associated bounding boxes, the co-occurrence matrix of patterns is generated based on street-style images.

Three state-of-the-art neural networks for object detection are evaluated by training on 45K street-style images (Section \ref{sec:DatasetLoc}). Faster-R-CNN (Faster Region-based Convolutional Neural Network) \cite{ren15} is the first quasi-real-time network. Its architecture is based on a Region Proposal Network (RPN) for quick region proposals and a classifier network for assessing the proposals. A non-maximum suppression (NMS) algorithm suppresses the boxes that are redundant. These steps are provided by the same base network, saving computational time. SSD (Single Shot Multi-Box Detector) \cite{liu15SSD} is the best network for optimising speed at the cost of a small drop in accuracy. The structure is equivalent to a number of class-specific RPN working at different scales. The results are then combined by NMS. R-FCN (Region-based Fully Convolutional Networks) \cite{dai16rfcn} is yet another improvement of Faster-R-CNN with a structure based on an RPN and a classifier. It has a reduced overhead due to the reduction in the size of the fully connected classifier that facilitates the classification of the different regions of the proposal independently. 

\subsection*{Recommendation architecture} 
\label{sec:recommendation}

In the next section, we describe three different methods for recommending items to users that are based on visual content such as the dominant colour, texture pattern, etc.

\subsubsection*{Recommendation by colour} \label{sec:colour}

A common attribute that encompasses consumer behaviour is to recommend garments based on colour. We operationalize such a scheme by (a) segmenting clothing items from street-style images, (b) extracting the dominant colour from segmented items using density estimation, (c) finding the associations of segmented items in the street-style dataset and (d) building a joint distribution of associations based on co-occurrence of dominant colours, from street-style images.

\begin{figure}
 \centering \includegraphics[width=0.5\textwidth, height=2.5in]
 {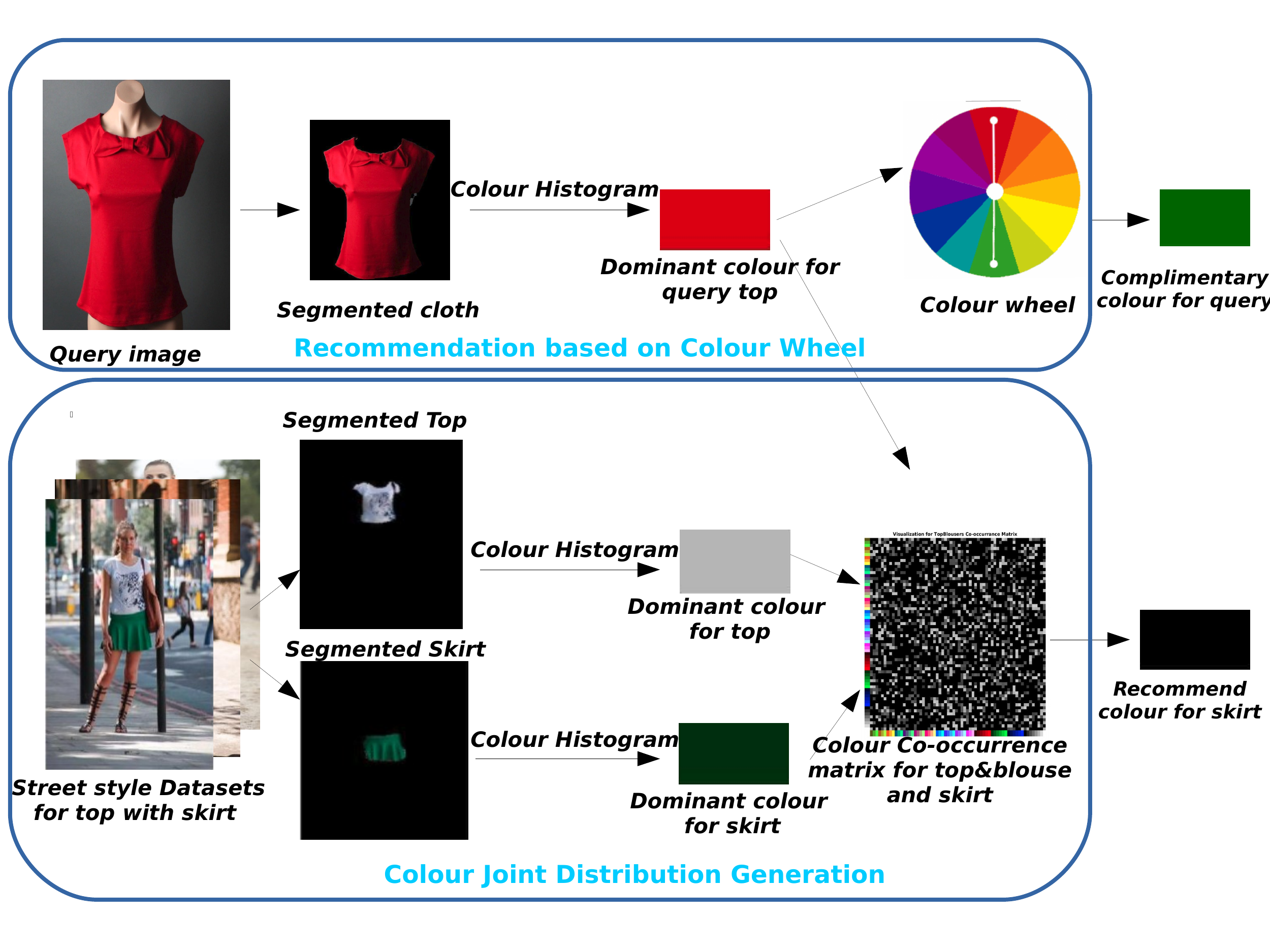}
 \caption{Framework on colour-based FRS. To generate colour co-occurrence matrix (bottom), we segment garments from street-style images and then find the dominant colour by calculating colour histogram from segmented pixels. After an association is generated, a joint distribution of colour between the top and bottom pairs (i.e., top with a skirt) is computed across the street-style dataset. When submitting a query top, a dominant colour is extracted from the segmented pixels. By searching the corresponding co-occurrence matrix, the best match colour of bottom garment is recommended. When determining dominant colour of query, the system can also recommend items with a specific attributes (i.e. complementary colour; top right).}
\label{fig_Framework_ColourFRS}
\end{figure}

First of all, a colour map is generated by using k-means based on CLElab \cite{ICE1977} value of segmented pixels in individual garment category. Each category has its own colour map; these maps are then used as an index of the co-occurrence matrix.  When a query is submitted, the dominant colour is extracted from the segmented garment and a search is initiated on the corresponding co-occurrence matrix to find the colour that best matches the query garment. Finally, we recommend inventory images from the corresponding category that has the matching colour to go with the query image. Figure \ref{fig_Framework_ColourFRS} shows the framework for dominant colour extraction and co-occurrence matrix generation from street-style image dataset. Additionally,  when the dominant colour is extracted from the query image, we use a knowledge-based recommendation engine wherein a colour wheel is also utilised to recommend items with specific attributes -- for example, complementary colour, triadic colour, etc. \cite{ColourCloth}.  

\subsubsection*{Recommendation by pattern} \label{sec:pattern}

\begin{figure}
 \centering \includegraphics[width=0.5\textwidth, height=2.5in]
 {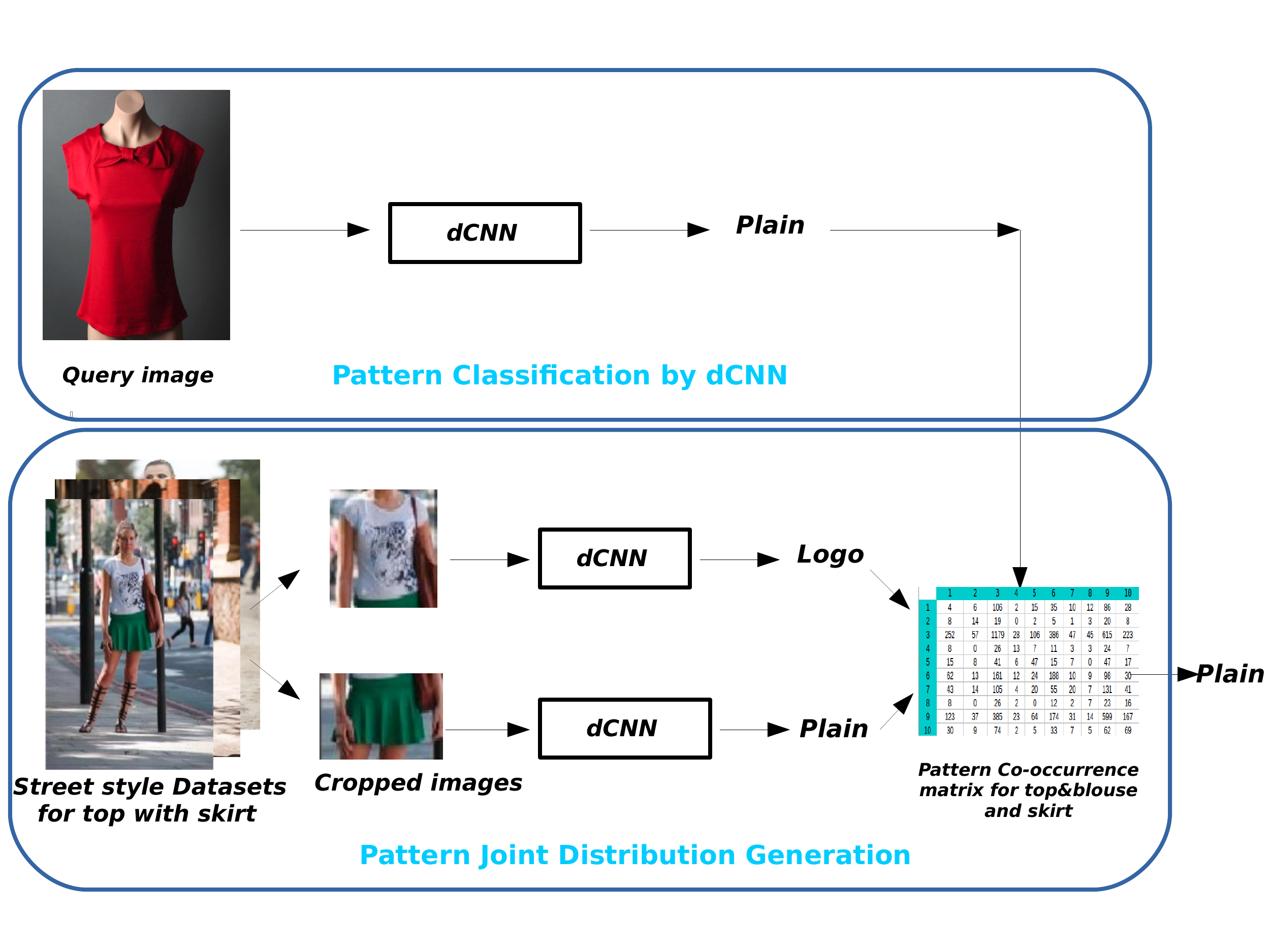}
 \caption{Framework on pattern-based FRS: Garments are cropped from street-style images. Each cropped garment is classified to one of the texture pattern category. A joint distribution of all patterns is generated based on pattern association (top and bottom garment) from street-style images. As a query is submitted, we classify it to one of the texture pattern, and return the best match pattern.}
\label{fig_Framework_PatternFRS}
\end{figure}

A similar framework for pattern recommendation, as shown in Figure \ref{fig_Framework_PatternFRS} is used to make a recommendation based on the pattern that is intrinsic to an object. Again, we recommend items that have a similar pattern by (a) detecting garments from street-style images, (b) classifying cropped garments to one of the texture patterns and (c) searching corresponding co-occurrence of texture pattern from street-style images. 

\subsubsection*{Recommendation via content-based retrieval} \label{sec:feedback}

\begin{figure}
\centering \includegraphics[width=0.5\textwidth, height=2.5in]
{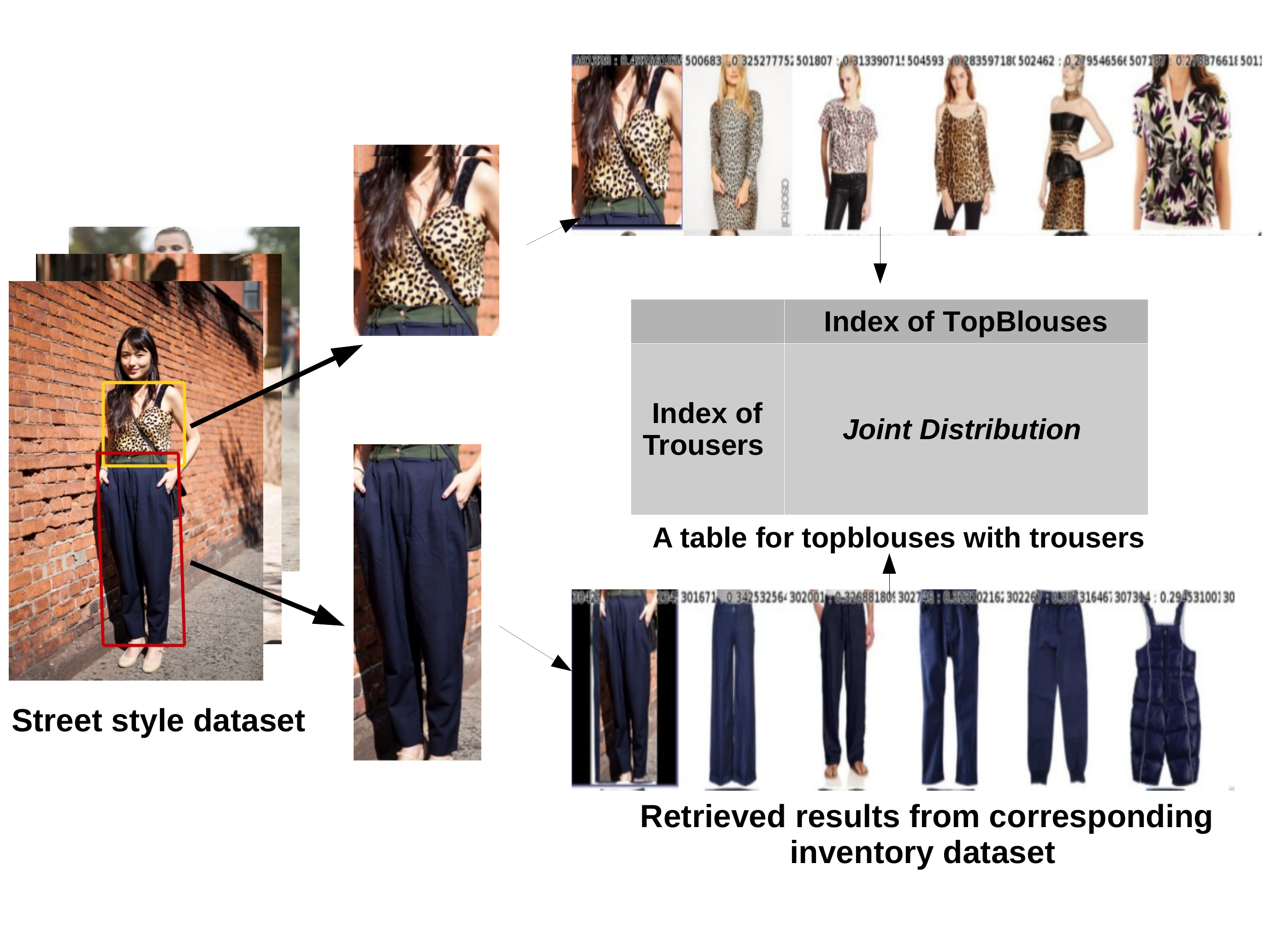}
\caption{Framework on retrieval-based FRS: Garments are first cropped from street-style images. As query image, each cropped garment is ingested in the retrieval system to find similar items from inventory images. By calculating association of a retrieved top and bottom items, a joint distribution is generated from the inventory dataset. Given a query top, a bottom garment may then be recommended by using retrieval on the query image and then looking up in the co-occurrence table.}
\label{fig_Framework_retrievalFRS}
\end{figure}

In Figure \ref{fig_Framework_retrievalFRS}, a content-based recommendation system is operationalised as follows: (a) locate and crop garments in street-style images, (b) associate top and bottom garments worn by the same person, (c) generate a table from each associated top-bottom pair of the inventory dataset. Specifically, we initiate a query on the top cropped garment from street-style image against inventory images of the same category (e.g., query a street-style blouse against inventory blouses), similarly, run a query on the bottom garment, for details on the specific architecture used for retrieval please see \cite{Sengupta2017}. A joint table can be constructed by adding the score of all possible combinations of top 5 retrieval results for the top garment (blouse) with top 5 retrieval results for bottom garment  (i.e., trouser). Such a table tells us how fashionable such a garment combination is. Given an image of a blouse, for example, a skirt may then be recommended by using retrieval on query image and then looking up in the table which of the skirts gather higher frequencies when combined with the blouse. 

\section{Datasets and Results} 
\label{sec:datasets and results}

\subsection*{Fashion Datasets} 
\label{sec:datasets}

\begin{figure}
 \centering \includegraphics[width=0.5\textwidth, height=2.5in]
 {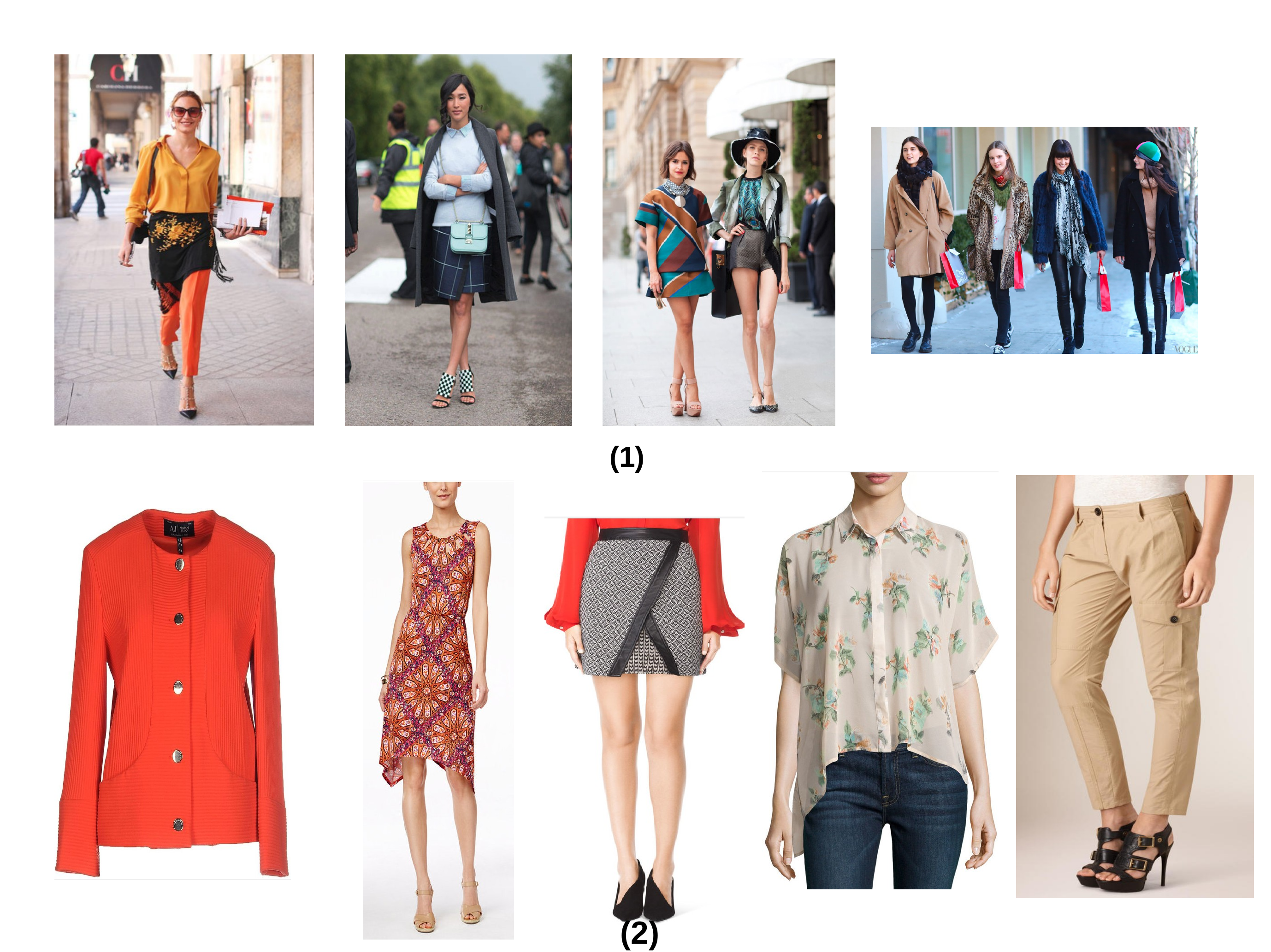}
 \caption{Two fashion datasets for recommendation: (1) street-style images (2) inventory images.}
 \label{fig_FashionDatasets}
\end{figure}

In order to recommend inventory images based on fashion trends
(street-to-shop), we generate two fashion datasets i.e., a street-style image dataset (no. 1) and an inventory image dataset (no. 2; Figure \ref{fig_FashionDatasets}). Dataset no.1 has 280K street-style images that were downloaded from latest fashion blogs. Out of these, 70K street-style images were used to build co-occurrence matrices of fashion inventories. Data set no. 2 has 100K inventory images that come from various fashion retailers. The images are categorised according to 5 classes, i.e., Coat/Jackets, Dresses, Skirts, Top/Blouses and Trousers. Inventory images are recommended to users based on colour, pattern and visual similarity by querying the generated co-occurrence matrices from street-style images. As seen in Figure \ref{fig_FashionDatasets}, most street-style images contain cluttered backgrounds: a person with a larger pose variation, multiple persons who overlap -- stand side by side, etc. Due to such backgrounds, object detection and segmentation is applied for street-style images to localise the requisite fashion items. Most inventory images show a single item with a plain background; localisation is also required here to mask out the model's leg and her head. 

\subsubsection*{Dataset for segmentation} \label{sec:DatasetSeg}

10K street-style images are generated and manually segmented by using the GrabCut algorithm \cite{Rother04GrabCut}. The dataset is split by using 7K images for training, 1.5K for validation and 1.5K for testing. Table \ref{table:Segmentation datasets} shows the split results for training, testing and validation dataset with a number of masks for each fashion item.    

\begin{table*}[ht]
  \begin{center}
    \begin{tabular}{cccccc}
        \toprule
        Dataset  & CoatsJackets & Dresses & Skirts & TopsBlouses & Trousers \\
        \midrule 
        Train(7K images)   & 2507 masks(22\%) & 2080(19\%) & 2018(18\%) & 2637(23\%) & 2078(18\%) \\ 
         \midrule 
        Validation(1.5K) & 557(23\%) & 470(19\%) & 419(17\%) & 542(22\%) & 426(19\%) \\ 
         \midrule
        Test(1.5K)   & 529(22\%) & 444(18\%) & 431(18\%) & 590(24\%) & 426(18\%) \\  
        \midrule 
        All Images(10K) & 3593(22\%) & 2994(19\%) & 2868(18\%) & 3769(23\%) & 2951(18\%) \\  
         \bottomrule
    \end{tabular}
  \end{center}
  \caption{Datasets for segmentation: each row shows the split for training, validation and test datasets, i.e., the total number of images in each dataset and masks (distribution of masks) in each category.}
  \label{table:Segmentation datasets}
\end{table*}

\subsubsection*{Dataset for localisation} 
\label{sec:DatasetLoc}
45K street-style images are generated and annotated manually by drawing a bounding box around each fashion item. The data is split into  36K images for training, 4.5K for validation and 4.5K for testing. The split results for training, testing and validation dataset with bounding boxes (BBs) for each item is shown in Table \ref{table:localizationdatasets}.

\begin{table*}[ht]
  \begin{center}
    \begin{tabular}{cccccc}
        \toprule
        Dataset  & CoatsJackets & Dresses & Skirts & TopsBlouses & Trousers \\
        \midrule 
        Train(36516 images)        & 16042(BBs)  & 7879  & 13257  & 25931  & 11584 \\ 
        \midrule 
        Validation(4565)    & 1985   & 1026  & 1691   & 3292   & 1429 \\ 
        \midrule
        Test(4564)          & 1930   & 980   & 1686   & 3257   & 1412 \\
        \midrule 
        All Images(45645)   & 19957  & 9885  & 16634  & 32480  & 14425 \\  
        \bottomrule
    \end{tabular}
  \end{center}
  \caption{Datasets for localisation: the  table lists the total number of bounding box for each category in  different datasets. }
  \label{table:localizationdatasets}
\end{table*}

\subsubsection*{Dataset for texture classification} 
\label{sec:DatasetCla}
\begin{figure}
 \centering \includegraphics[width=0.4\textwidth, height=1.75in]
 {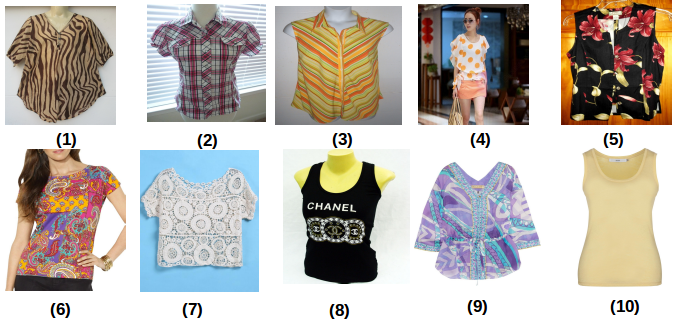}
 \caption{Pattern dataset: (1) animal print (2) checks (3) stripes (4) dots (5) floral (6) paisley (7) crochet (8) logo (9) cosmic (10) plain.}
\label{fig_PatternDataset}
\end{figure}

Combining some categories in DTD \cite{cimpoi14describing}, texture tags in Deep Fashion dataset \cite{Liu2016} and some fashion blogs,  10 most popular pattern categories on fashion inventory are selected. 14K single fashion item images (11K for training and 3K for testing) from a client is sourced for training a neural network model to classify the pattern behind each texture; some examples for the 10 categories are shown in Figure \ref{fig_PatternDataset}.

\begin{table*}[ht]
  \begin{center}
    \begin{tabular}{cccccccccc}
        \toprule
        Dataset  & AnimalPrint & Check & Stripes & Dots & Floral  & Crochet  & Logo & Cosmic & Plain \\  
        \midrule 
        Train     & 979(images)  & 1142   & 2587   & 975  & 2011   & 307   & 407   & 468 & 1836  \\   
        \midrule 
        Test     & 246  & 286   & 648   & 244  & 503   & 76   & 102   & 117 & 460  \\             
        \bottomrule
    \end{tabular}
  \end{center}
  \caption{Pattern dataset for classification: image split results for each category for train and test datasets.}
  \label{table:PattenDataset}
\end{table*}
 
\subsection*{Results} 
\label{sec:Results}

\subsubsection*{Results for segmentation} 
\label{sec:ResultsSeg}

\begin{figure}
 \centering \includegraphics[width=0.5\textwidth, height=2.25in]
 {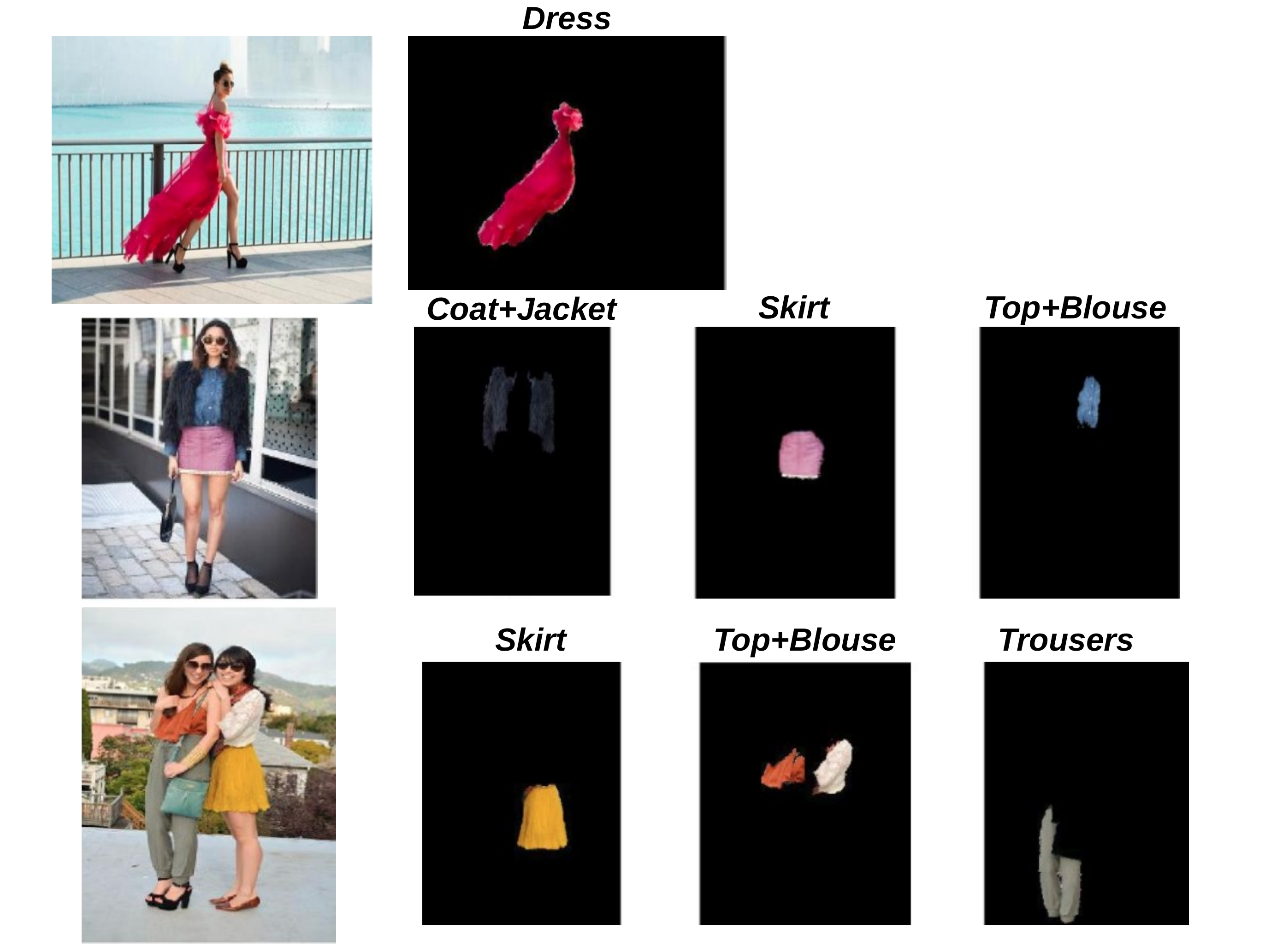}
 \caption{Examples for segmentation: first row shows segmented dress; second row is segmentation results for multiple garments, third row is segmented garments from multiple persons  within one image.}
 \label{fig_SegmentationExamples}
\end{figure}

Two methods for segmenting objects (Section \ref{sec:segmentation}) are evaluated on street-style images. Both models are initialized with a pretrained VGG-16 model on ImageNet, trained on 8.5K (train+validation) segmentation dataset and evaluated on a 1.5K testing dataset. Table \ref{table:Segmenation evaluation} shows Intersection over Union (IoU = True positive/(True positive+False positive+False Negative)) and Pixel Accuracy (PA = True positive/(True positive+False Negative)) on each fashion class and the mean value of each class for both the models. DeepLab with multiple scales and LargeFOV along with a CRF achieves the best performance of 59.66\% mean IoU and 73.99\% mean PA. It is also evident that combining CRF with FCN increases the  Mean IoU by 4\% and PA by 2\%. We have also trained the model by initializing with VGG-16 on our fashion classification dataset \cite{Sengupta2017}; here, the mean IoU drops by  4\%. Figure \ref{fig_SegmentationExamples} shows  segmentation results on street-style images by using the DeepLab-MultipleScales-LargeFOV algorithmic combination with a CRF. Average test time to segment an image: FCN = 115ms and CRF=638ms on a nVIDIA Titan X GPU with 12GB memory.           

\begin{table*}[ht]
  \begin{center}
    \begin{tabular}{ccccccccc}
        \toprule
        Models  & CoatsJackets & Dresses & Skirts & TopsBlouses & Trousers & Background & Mean IoU & Mean PA\\
        \midrule 
        FCN-8   & IoU = 52.74\% & 74.23\% & 64.72\% & 57.91\% & 71.85\% & 96.36\% & 69.63\% & \\ 
                & PA = 44.80\%  & 44.73\% & 42.88\% & 36.11\% & 54.38\% & 94.07\% &   &52.83\%\\
        \midrule 
        FCN-8+CRF & 55.93\% & 76.50\% & 65.89\% & 58.94\% & 75.35\% & 97.10\% & 71.62\% &\\ 
                  & 48.24\%  & 48.94\% & 47.09\% & 40.69\% & 58.77\% & 94.79\% & &56.42\%\\
        \midrule
        DeepLab-MSc-LargeFOV   & 69.39\% & 67.69\% & 65.69\% & 55.63\% & 78.33\% & 96.85\% & 72.28\% &\\ 
                  & 51.46\%  & 48.03\% & 46.89\% & 39.18\% & 56.81\% & 94.67\% & &56.17\%\\
        \midrule  
        DeepLab-MSc-LargeFOV + CRF & 70.70\% & 70.94\% & 68.69\% & 56.14\% & 80.01\% & 97.45\% & 73.99\% &\\ 
                  & 55.12\%  & 52.50\% & 51.70\% & 42.31\% & 61.14\% & 95.20\% & &59.66\%\\
        \bottomrule
    \end{tabular}
  \end{center}
  \caption{Evaluation on segmentation: each row for individual methods show IoU (top line) and PA (bottom line) values in different categories. MSc refers to multiple scales.}
  \label{table:Segmenation evaluation}
\end{table*}

\subsubsection*{Results for localisation} 
\label{sec:ResultsLoc}

\begin{figure}
 \centering \includegraphics[width=0.4\textwidth, height=2in]
 {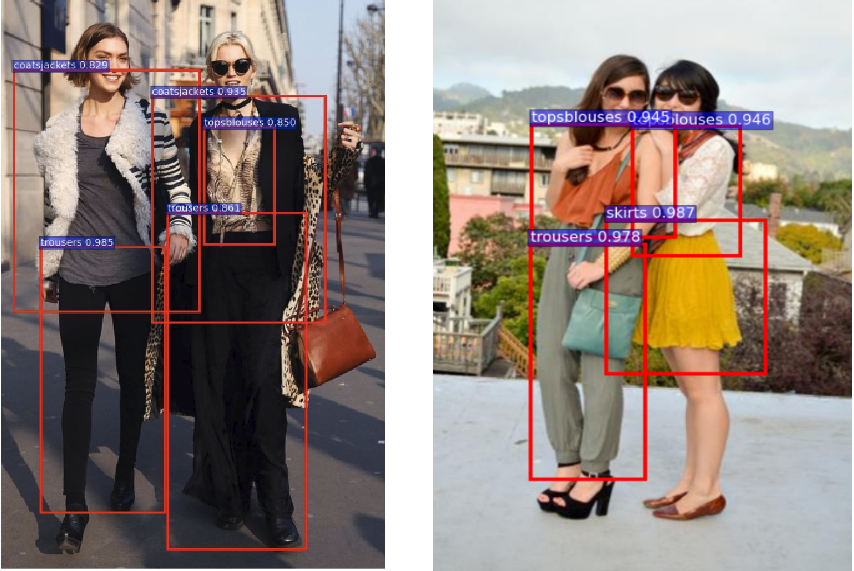}
 \caption{Examples for localisation}
 \label{fig_LocalizationExamples}
\end{figure}

In order to evaluate the performance of three networks (Section \ref{sec:localization}) on localisation, three models are trained on 40.5K (train+validation) street-style localisation dataset and tested on a 4.5K dataset. We used the default parameters chosen from the original papers. Table \ref{table:localization evaluation} shows the Average Precision (AP) calculated on each object class and the mean Average Precisions (mAPs) for the three models. The bounding boxes are considered only if the IoU is larger than 50\%. Average testing time for an image is evaluated on a NVIDIA Quadro M6000 GPU with 24GB memory. Table \ref{table:localization evaluation} shows that R-FCN has an edge over the other models that were evaluated. SSD is particularly suitable when speed is the main concern. Figure \ref{fig_LocalizationExamples} shows  R-FCN detection results on street-style images.    

\begin{table*}[ht]
  \begin{center}
    \begin{tabular}{cccccccc}
        \toprule
        Models  & CoatsJackets & Dresses & Skirts & TopsBlouses & Trousers  & mAP  & Time(ms)\\  
        \midrule 
        R-CNN   & AP=83\%        & 73\%    & 75\%   & 69\%     & 84\%   & 76.8\%   & 141\\                
        \midrule 
        SSD 500x500  & 84\%         & 74\%  & 76\%   & 76\%        & 86\%       & 79.2 \% &70\\
        \midrule
        R-FCN   & 89\%         & 78\%  & 81\%   & 78\%        & 91\%       & 83.4 \% &137\\
        \bottomrule
    \end{tabular}
  \end{center}
  \caption{Evaluation on localisation:  for three models, each row shows the Average Precision (AP) on each class and the mean Average Precisions (mAPs). Run times for each method is also shown. The bounding boxes are considered only if the IoU is larger than 50\%}
  \label{table:localization evaluation}
\end{table*}

\subsubsection*{Results for texture classification} \label{sec:ResultsCla}
For the pattern recommendation system, the cropped garments are classified in 10 texture patterns. For this,  a pattern classifier is  trained by fine tuning a pre-trained VGG-16. We use 11K images for training and  3K images for testing (Section \ref{sec:DatasetCla}). Results are listed in Table \ref{table:PattenClassification evaluation}. 

\begin{table*}[ht]
  \begin{center}
    \begin{tabular}{ccccccccccc}
        \toprule
        Models  & AnimalPrint & Check & Stripes & Dots & Floral  & Crochet  & Logo & Cosmic & Plain & Mean Accuracy \\  
        \midrule 
        VGG-16     & 77.6\%  & 87.4\%   & 95.5\%   & 84.8\%  & 90.9\%   & 72.7\%   & 86.3\%   & 70.1\% & 96.7\% & 88.9\% \\                
        \bottomrule
    \end{tabular}
  \end{center}
  \caption{Evaluation on pattern classification}
  \label{table:PattenClassification evaluation}
\end{table*}

\subsubsection*{Results for garment association} \label{sec:ResultsLoc}
After detecting garments, a  person detector \cite{Dollar2010} is applied to constrain the cropped garment being worn by the same person. In total, 6 associations between a pair of garments in 70K street-style images are generated and listed in Table \ref{table:Association}. The numbers indicate how many people wear corresponding garments in the 70K street-style images. 

\begin{table}
  \begin{center}
    \begin{tabular}{ccccc}
        \toprule
        Category       & Dresses    & Skirts     & Trousers  & TopsBlouses\\
        \midrule 
        CoatsJackets   & 3744       & 4001       & 7393      & 5609  \\ 
         \midrule 
        TopsBlouses    & null       & 5376       & 6058      & null\\ 
         \bottomrule
    \end{tabular}
  \end{center}
  \caption{Association pairs on 70K streetstyle images}
  \label{table:Association}
\end{table}

\subsubsection*{Recommendation using colour} \label{sec:RecCol}
\begin{figure*}[ht]
 \centering \includegraphics[width=0.7\textwidth, height=4in]
 {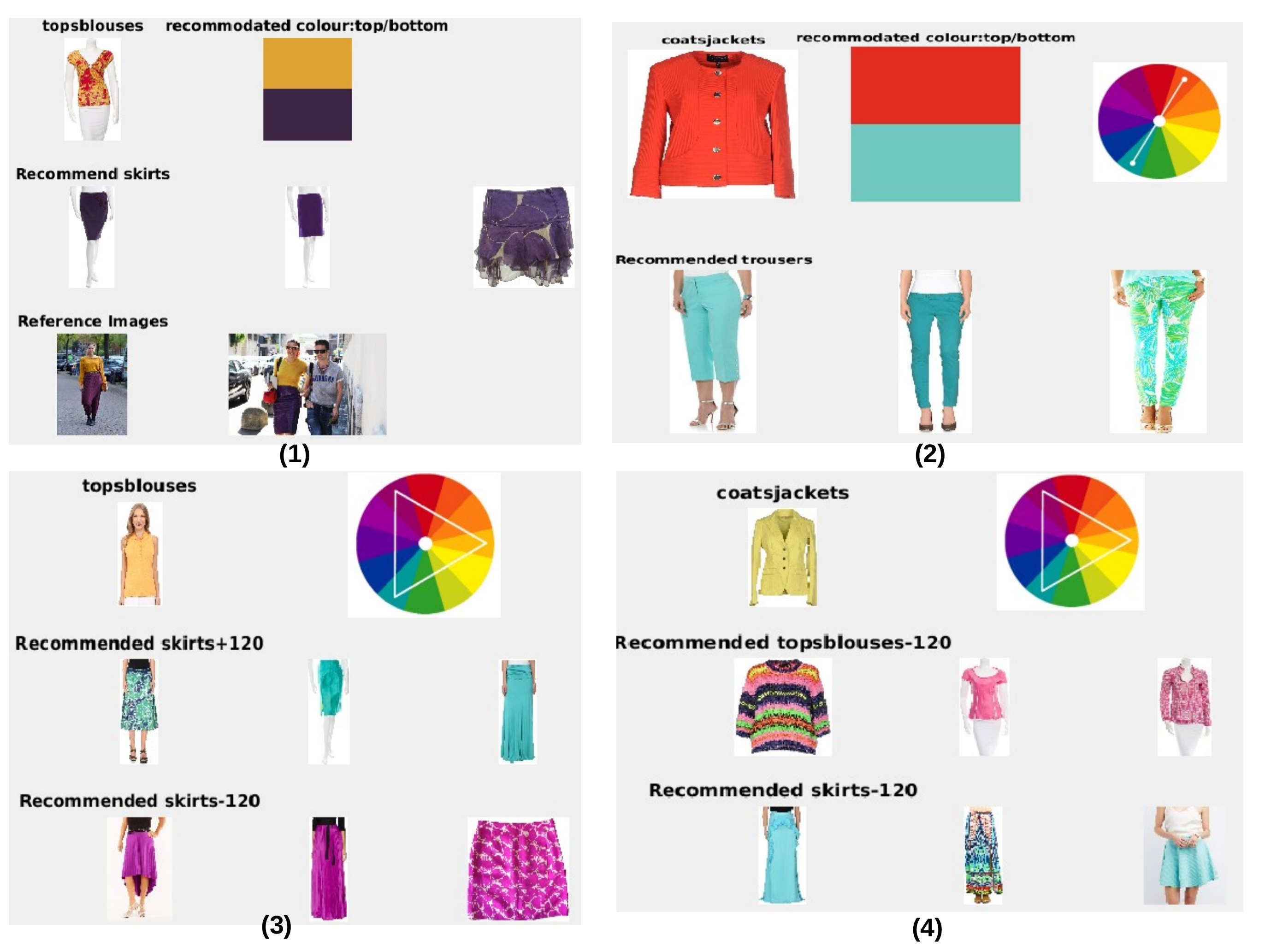}
 \caption{Examples of colour-based FRS: (1) FRS on colour occurrence matrix: first row  displays the dominant colour of query top (shown in left) and recommended colour of bottom garment (i.e., skirt). The recommended garments from inventory images are shown in second row. The third row provides the same matching items from street-style dataset as reference. (2)-(4) The recommended colour is  based on fashion colour rules: (2) FRS on complementary colour,  (3) FRS on one of the triadic colour and (4) FRS on triadic colour for three different fashion item associations.}
 \label{fig_FRS_Colour}
\end{figure*}

After garment association, 6 co-occurrence matrices of dominant colour are generated from 70K street-style images. In our system, a colour map with 130 bins for each category is created by using k-means on all segmented pixels of the corresponding garment in street-style images.  When a query image is submitted, the dominant colour is extracted from the segmented item and a search from corresponding co-occurrence matrix is initiated to find the best colour that matches the query item. For example in Figure \ref{fig_FRS_Colour}(1), the first row shows the query image and best matching colour obtained from the tops/blouses-skirts colour co-concurrence matrix. The second row shows the recommend skirt according to the recommended colour from an inventory database; some reference examples with same match colour from the street-style dataset are displayed in the third row. In Figure \ref{fig_FRS_Colour}(2-4) we show some examples for the colour recommendation based on different aspects of the colour wheel. Figure \ref{fig_FRS_Colour}(2) shows complimentary colour trousers with a query top. Figure \ref{fig_FRS_Colour}(3) shows one of the triadic coloured skirts with a yellow top. Figure \ref{fig_FRS_Colour}(4) shows triadic coloured skirts and tops/blouses with a yellow coat.         

\subsubsection*{Recommendation using pattern} \label{sec:RecPat}

\begin{figure*}[ht]
 \centering \includegraphics[scale=0.4]
 {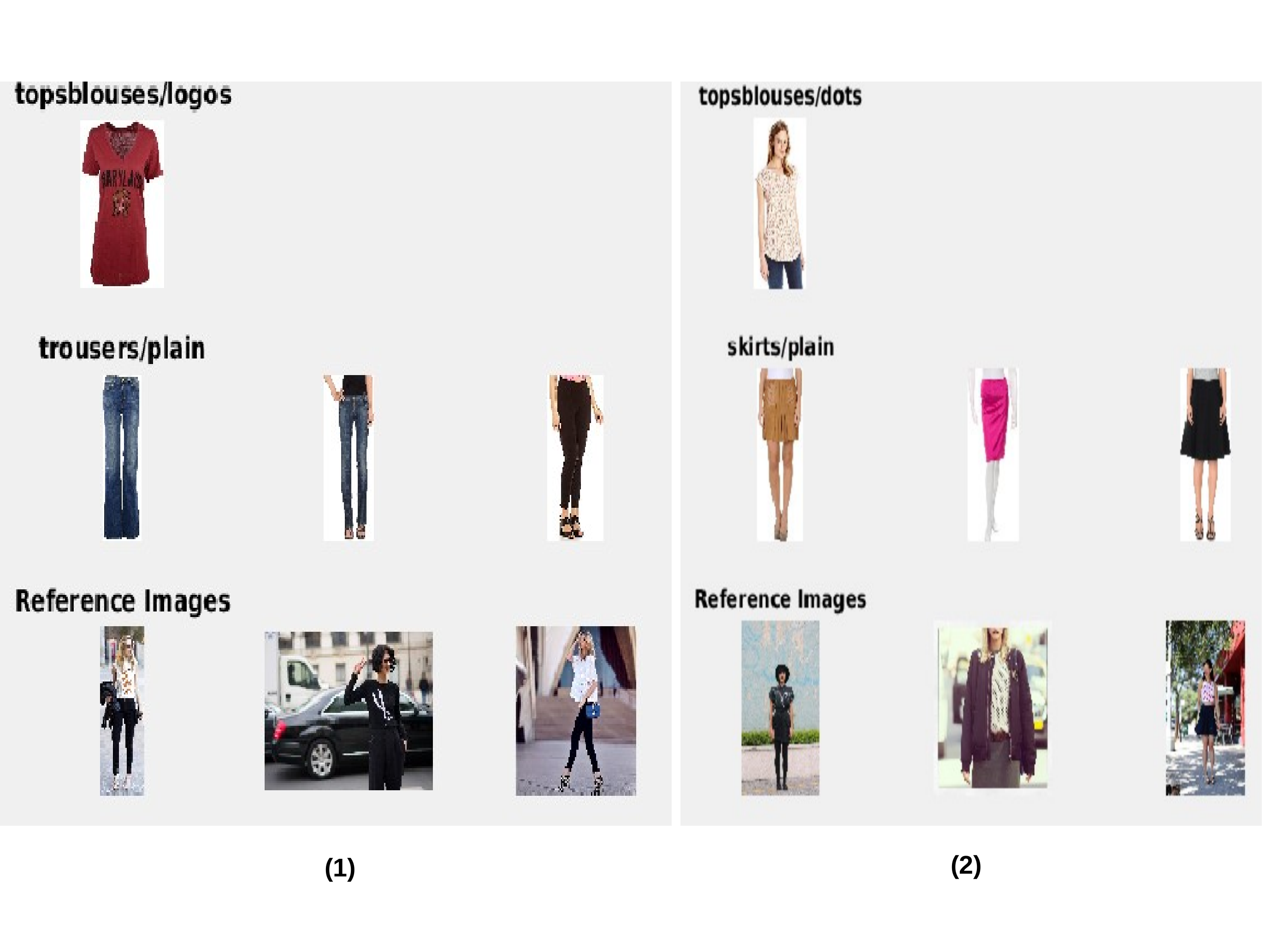}
 \caption{Examples of texture pattern-based FRS. (1) top/blouses with trousers. (2) top/blouses with skirts}
 \label{fig_FRS_Pattern}
\end{figure*}

For pattern recommendation, when a query image is submitted, the garments are cropped from the images and classified in to one of the ten texture patterns. We then search a corresponding $10$ x $10$ pattern co-occurrence matrix to find the best match pattern with respect to the query item. This then allows us to recommend items with a matching pattern from the inventory dataset. Two examples are shown in Figure \ref{fig_FRS_Pattern}: (1) shows the top/blouse that form the query; a plain colour trouser is recommended to take into account the attributes of the query pattern. Figure \ref{fig_FRS_Pattern}(2) shows the query i.e., top/blouse with a dotted pattern;  the FRS recommends that a plain coloured skirt is worn with such a top. The third row in each figure shows some reference images from the street-style dataset with the same match pattern. 

\subsubsection*{Recommendation using content-based retrieval} 
\label{sec:RecImp}

\begin{figure*}[ht]
 \centering \includegraphics[scale=0.3]
 {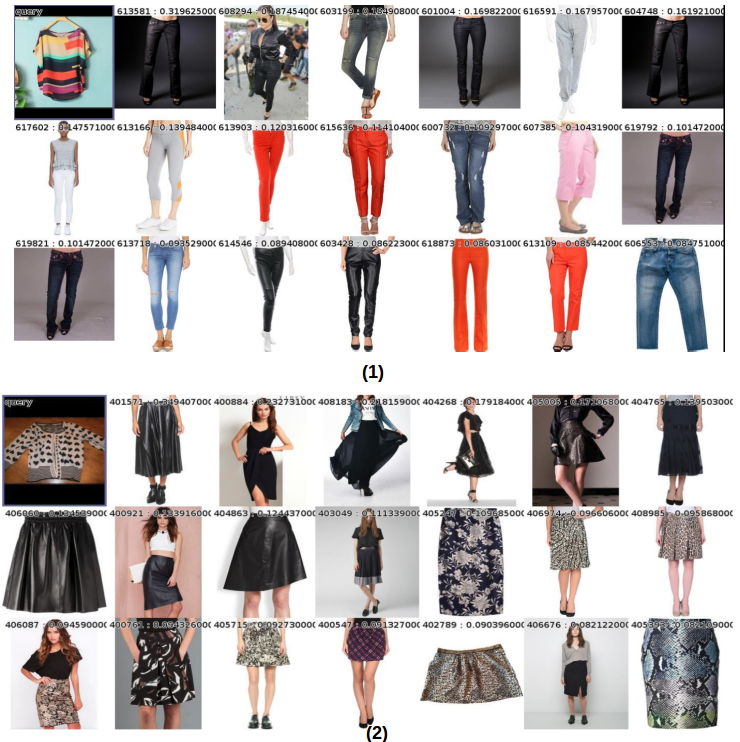}
 \caption{Examples of retrieval-based FRS:(1) top/blouses with trousers and (2) coat/jackets with skirts. The recommended images from inventory images are ranked in decreasing order of joint frequency.}
 \label{fig_DataDrived_FRS}
\end{figure*}

Given a query image, we run a query on the image against inventory images of the same ``top" category \cite{Qian2017},  pick some of higher ranking ``top" garments, use the look-up table to find and recommend the most frequent ``bottom" garments for each ``top" garment to the user. Two examples are shown in Figure \ref{fig_DataDrived_FRS}: (1) top/blouses with trousers and (2) coat/jackets with skirts.

\section{Discussion}
\label{sec:discuss}

This paper has detailed an end-to-end commercially deployable system -- starting from image segmentation, localisation to recommending a dyad of clothing accessories. The knowledge representation is learnt by crawling through fashion blogs (street-style oracles) for images that are prescriptive of a variety of style, preferred by consumers. Deep neural networks complement this knowledge by learning a latent feature representation which then enables dyadic recommendations. We propose two other simpler recommendations by utilising the colour wheel to prescribe dyads of colours or use deep feature vectors to recommend clothing accessories based on the texture of the fabric. The framework is scalable and has been deployed on cloud-service providers. 

Our work adds on to the burgeoning vertical of algorithmic clothing \cite{Liu2016} that use discriminative and probabilistic models to recommend consumers on how to finesse their dressing style. For example, \cite{Murillo2012} has learnt `urban tribes' by learning which group of people are more likely to socialise with one another, therefore may have similar dressing style. Classifying styles of clothing has been the focus of \cite{Bossard2013}'s work where the authors use a random forest classifier to distinguish a variety of dressing styles. Similarly, \cite{Oramas2016} use neural networks to measure the visual compatibility of different clothing item by analysing co-occurrences of base-level elements between images of compatible objects. The combination of street-style oracles with a deep learning based feature representation framework is similar to the work by \cite{Veit2015} wherein they use a Siamese convolutional neural network to learn compatibility of a variety of clothing items. There is a stark dissimilarity though -- \cite{Veit2015} used \url{Amazon.com}'s co-purchase dataset, which is instantiated on the assumption that two items purchased together are worn together. This may not be always true -- therefore, the present work bases recommendation on the current trends in fashion (well represented by fashion blogs). The framework is flexible such that `clothing trends' can be updated to keep up with seasonality trends \cite{Al-Halah2017} or dissected into hierarchial models suited to the demographics or age-range of the clientele. With the availability of GPUs, such a framework becomes highly scalable.

There are a few challenges that can imperil the formation of a joint occurrence matrix. The first is the sparsity of the matrix involved -- this is caused due to an inadequate number of street-style images with a specific combination of two clothing items. An easy way to alleviate such an issue is to use generative models \cite{Murphy2012}. A neural network can also be utilised as a function approximator (see below) such that the learnt features can be encoded \cite{Sengupta2017a} to reveal dependencies between inventory items. 

Our framework would recommend similar items to those previously suggested to the user. Whilst such a problem is severe for collaborative filtering approaches,  our hybrid recommender system alleviates just a part of it, especially if we relax the assumption that the co-occurrence matrix has quite a stable probability distribution. Thus, a vital strand of our current research lies in personalization -- how can we alter the recommendations such that it takes into account not only our shopping behaviour but also the granularity of our `personal' taste. One way forward to formulate this feature-based exploration/exploitation problem is to frame it as a contextual bandit problem \cite{Langford2008a}. Put simply, such an algorithm sequentially selects the dyad recommendation based on the interaction of the consumer with the recommendation system.   

The present work focuses on a recommendation dyad i.e., a trouser to go with a shirt; nevertheless, the present framework is equipped to make recommendations over a much larger combination of co-occurrences. As earlier, the next step forward would be to replace the joint-occurrence matrix with a neural network so that a non-linear function over multiple items could be learnt. This would be necessary for next generation algorithms that can recommend us an entire wardrobe rather than dyads or triads of clothing items. 

\section{Acknowledgment}
\label{sec:acknowledgment}
This work was supported by two Technology Strategy Board (TSB) UK grants (Ref: 720499 and Ref: 720695).

\bibliography{cortexica_recommend}
\bibliographystyle{ACM-Reference-Format}


\end{document}